\documentclass[letterpaper, 10 pt, conference]{IEEEtran}
\IEEEoverridecommandlockouts
\usepackage{algorithmic}
\usepackage{mathtools}
\usepackage{amsmath}
\usepackage{amsfonts}
\usepackage{cite}
\usepackage{hyperref}
\usepackage{subcaption}
\usepackage{listings}

\title{\LARGE \bf \vspace{2.0cm} Elevation State-Space: Surfel-Based Navigation in
Uneven Environments for Mobile Robots}
\author{
    Fetullah Atas $^{1}$  Grzegorz Cielniak  $^{2}$ Lars Grimstad $^{1}$
	\thanks{$^{1}$ Norwegian University of Life Sciences {\tt\small \{fetullah.atas, lars.grimstad\} @nmbu.no}}
	\thanks{$^{2}$ University of Lincoln {\tt\small gcielniak@lincoln.ac.uk}}
}

\begin{document}

\maketitle
\thispagestyle{empty}
\pagestyle{empty}

\begin{abstract}
This paper introduces a new method for robot motion planning and navigation in uneven environments through a surfel representation of underlying point clouds. The proposed method addresses the shortcomings of state-of-the-art navigation methods by incorporating both kinematic and physical constraints of a robot with standard motion planning algorithms (e.g., those from the Open Motion Planning Library), thus enabling efficient sampling-based planners for challenging uneven terrain navigation on raw point cloud maps. Unlike techniques based on Digital Elevation Maps (DEMs), our novel surfel-based state-space formulation and implementation are based on raw point cloud maps, allowing for the modeling of overlapping surfaces such as bridges, piers, and tunnels. 
Experimental results demonstrate the robustness of the proposed method for robot navigation in real and simulated unstructured environments. The proposed approach also optimizes planners' performances by boosting their success rates up to 5x for challenging unstructured terrain planning and navigation, thanks to our surfel-based approach's robot constraint-aware sampling strategy.  
Finally, we provide an open-source implementation of the proposed method to benefit the robotics community.
\end{abstract}


\section{Introduction} \label{introduction}

There is an increasing demand for robots in uneven environments such as construction, rural, disaster, or planetary discovery sites. Current methods for autonomous navigation in these environments use various environment representations, including 2D grid maps~\cite{macenski2020marathon2, tbd}, Digital Elevation Maps (DEMs)~\cite{Wermelinger, STEP}, 3D occupancy grids~,\cite{WangChaoqun} and meshes~\cite{OVPC,mesh_nav}.

\begin{figure}
  \centering 
  \includegraphics[width=1\linewidth]{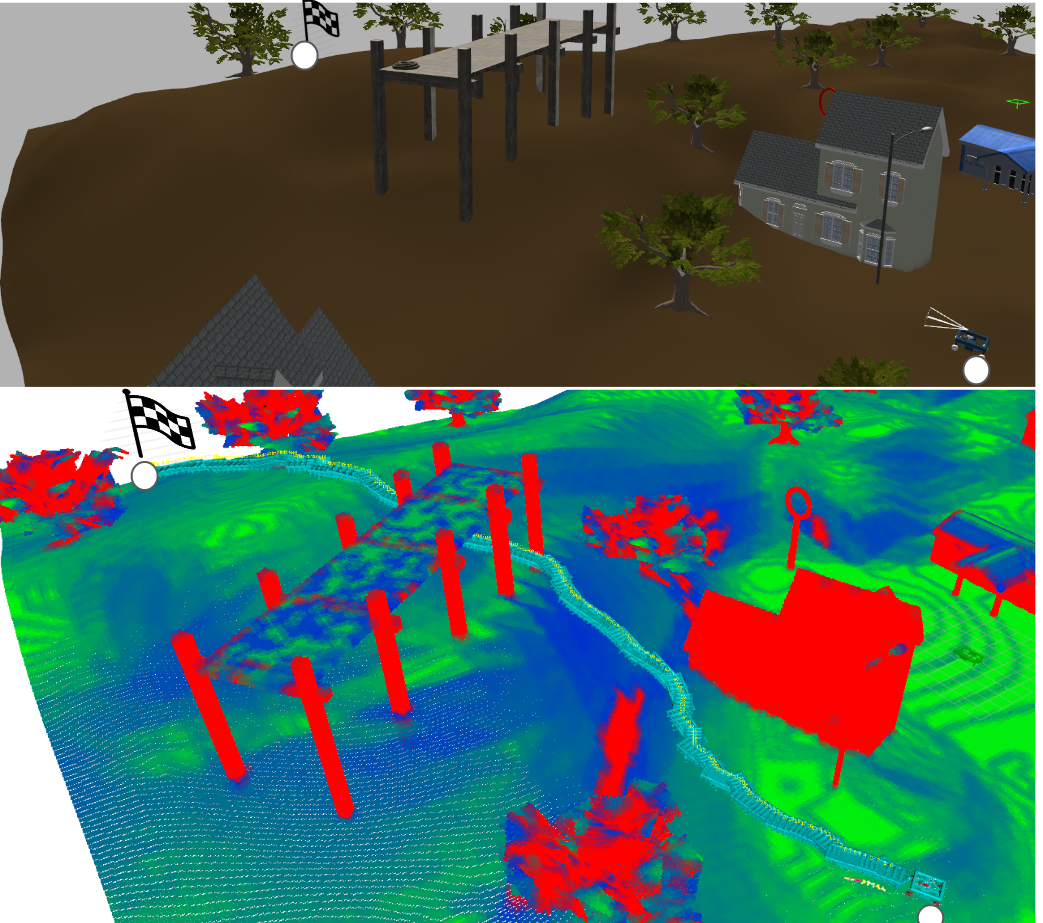}
  \caption{An example use-case for navigation in unstructured terrain. PRM* planner~\cite{karaman_frazolli},  operating in the proposed state-space, returns a valid path (in cyan color) by passing under the pier. }
  \label{fig:pier_teaser}
\end{figure}

Despite the variety of proposed methods, some limitations call for further improvements. A few examples in this regard are limitations from the environment representation (as in DEMs for not being able to model some structures such as tunnels, overhangs, etc.) and the curse of dimensionality in full 3D occupancy grids. There is also a lack of convergence towards a standard environment representation for 3D navigation. 2D grid maps are widely adopted for structured terrain planning and navigation in academia and industry. As we explore in Sec.~\ref{related_work}, there is no widely adopted representation for navigation in uneven terrains, leading to a need for alternative representations with improvements over existing methods. In earlier approaches, such as in~\cite{mesh_nav, Pfrunder}, the standard planning algorithms were tailored specifically to work with a particular environment representation requiring additional integration effort and limited portability between the representations. This caused a considerable effort to reimplement or adapt different planning algorithms to work with proposed environment representations.    
This work proposes a new state-space design and implementation to address these limitations. We built our method on top of raw point cloud maps. Point clouds are directly derived from distance measurements provided by various sensors (e.g., RGBD cameras, LIDARs, etc.). Point clouds are not volumetric structures that could convey the occupancy information of the environment. To convey the occupancy information of the environment, we create an \textit{occupancy volume} through surfels, which are introduced as point primitives to simplify the representation of a dense point cloud in computer graphics~\cite{surfels}. The surfels hold a \textit{cost} property which indicates a traversability measure of the region they represent within the dense point cloud. The cost property of surfels is also projected to \textit{occupancy volume}, see Fig.~\ref{fig:waffle_surfel}. The \textit{occupancy volume} is built where it matters: the traversable regions on the \textit{surface} of environments, which is advantageous since a ground robot's area of interest is primarily the ground's surface. We limit the occupancy volume used for planning to only relevant locations (ground's surface) without loss of ability to model complex structures such as tunnels or piers, which DEM-based methods cannot model, see Fig.~\ref{fig:pier_teaser}.
 Finally, together with the cost property, this volume is called \textit{elevation state-space}, a dedicated state-space for planning and navigation in unstructured terrains.
We run experiments in real and simulated unstructured environments to validate our approach. Through our surfel-based state-space implementation, we enable seamless integration with the standard planning techniques such as those from the OMPL~\cite{ompl} library for unstructured terrain planning and navigation.

In summary, our key contributions are: 

\begin{itemize}
    \item A configurable implementation interface that can account for different robot types with various kinematic and physical constraints, we run experiments on Ackerman and omnidirectional robot types.
    \item We demonstrate that both in real and simulated uneven terrains, the proposed approach leads to better results than standard navigation frameworks (i.e., Navigation2~\cite{macenski2020marathon2}).
    \item An open-source implementation of our software   \footnote{\url{https://github.com/NMBURobotics/vox_nav}} which enables the use of standard motion planning libraries (i.e., OMPL) for unstructured terrain planning and navigation, removing the need for additional efforts to integrate representation-dependant path planners. 
\end{itemize}

\section{Related Work} \label{related_work}

The previous work in mobile robot navigation in unstructured environments used a variety of map representations, including digital elevation maps (DEMs), 2D-3D occupancy grids, and meshes. We take a brief look into recent methods and explain their characteristics in the following.

A digital elevation map (DEM) is a generic grid map where each cell within the grid holds the elevation information of corresponding $x$ and $y$ locations with a function like $z$ = $f(x,y)$. Wermelinger et al.~\cite{Wermelinger} used a special type of DEM where, besides elevation, a traversability
measure of terrain was also present within each cell. The authors referred to the new form of this map as a traversability map. They used this map for planning paths for a legged robot in unstructured terrain. Similarly, Fan et al. \cite{STEP} used a layered grid map where the elevation and risk factors associated with each position were present. They then used this layered 2D grid map for risk-aware planning and navigation in unstructured environments. These methods rely on an elevation map with limitations such as not modeling overhangs, tunnels, etc. 

An OcTree-based 3D occupancy method, OctoMap, was introduced by Hornung et al.~\cite{octomap}. In OctoMap, occupancy values of geometric voxels are determined by a probabilistic approach to account for
sensor noise and uncertainties. Wang et al.~\cite{WangChaoqun} presented a framework that indirectly used OctoMap to navigate an
indoor mobile robot in an unstructured environment. First, an OctoMap of the domain is created. The generated OctoMap is then
projected to several layers with various resolutions, and differences of the gained layers are used to determine the traversability of the environment map. The planning and navigation are performed on top of this 2D traversability map. The method was developed and tested only for
indoor unstructured environments; hence there was no incorporation of terrain roughness or ground clearance of the robot chassis.

Some other methods, such as~,\cite{mesh_nav, OVPC} have used 3D meshes generated from 3D point cloud maps of the environment to represent unstructured terrains.  They used acquired 3D mesh edge connectivity to develop a navigation strategy for unstructured outdoor environments. Authors in~\cite{mesh_nav} use such a 3D mesh to build the edge connectivity of a 3D environment. Graph-based search algorithms are used on this edge connectivity to create feasible paths for navigation. Although the method is aimed to be robot-independent, there was no direct indication for consideration of kinematic constraints, such as Ackerman-type. Ruetz et al.~\cite{OVPC} also presented a 3D mesh-based framework. However, the acquired 3D mesh was only based on visible point clouds from the live LIDAR sensor, limiting the method for global navigation in unstructured environments.

Although initially not intended for unstructured terrain navigation, Navigation2~\cite{macenski2020marathon2} and its predecessor, the original ROS navigation framework~ is worth mentioning due to its popularity in academia and industry\cite{tbd}. These methods, which rely on a 2D occupancy grid for environment representation, are practical and have been used by numerous researchers. However, because these approaches are based on a 2D occupancy grid for environment representation, their use is limited for unstructured terrain.

There was less emphasis on the relationship between environment representation and planning in the existing methods, resulting in incompatible, duplicate implementations of path planning algorithms among used environment representations. In contrast, our approach places a greater emphasis on the relationship between environment representation and planning. We argue that seamless integration between the environment representation and planning is crucial for challenging unstructured terrain planning and navigation. Our approach integrates environment representation and planning tightly through our surfel-based elevation state-space, eliminating the need to tailor planners to a specific environment representation. As we show in Sec.~\ref{experiments}, this helps us navigate unstructured environments effortlessly as we enable access to dozens of efficient sampling-based planners (e.g., from OMPL).   

\section{Approach} \label{approach}


Surfel is an abbreviation for ``surface element''~\cite{surfels}, which is introduced as point primitives to simplify the rendering or shading of an object represented by a dense point cloud. For a ground robot navigating in an unstructured terrain where the environment is represented by such a dense point cloud, the area of interest is the surface of the point cloud map. We use these point primitives (surfels) to simplify the representation of terrain features derived from points, see Sec.~\ref{regressing_costs_subsection}. In the following, we describe the extraction of traversability information and embedding of this information into elevation state-space.

An environment is represented with point clouds consisting of $n$ points. We refer to this point set as the original point set $P_O$.
We perform a uniform sampling to discretize this environment and distribute the cost values associated with local patches \cite{pcl} operation on the original point cloud set $P_O$. Uniform sampling is performed with a voxel size of $d_s$, resulting in a point cloud subset of $j$ points. We refer to this point set as a sampled point set $P_S$. We extract local terrain features such as roughness, tilt, the maximum height of sampled points, and their geometrical relationship with neighboring points from the original point's set $P_O$. These features are used for deriving a traversability measure of the map, see the itemized cost critics in Sec.~\ref{regressing_costs_subsection}.

We construct the surfels with sampled points and their neighbors within radius $r_s$, from the \textit{original} point set $P_O$.

Let us refer to this acquired surfel set as the original surfel set $S_O$, each surfel in this set is deﬁned by a 3D position \begin{math} v_s \in \mathbb{R}^{3}\end{math}, a 3D normal vector \begin{math} n_s \in \mathbb{R}^{3} \end{math} and a radius \begin{math} r_s \in \mathbb{R}^{1} \end{math}. The radius of surfel \begin{math} r_s \end{math} is approximated to \begin{math} d_s/2.0 \end{math} to minimize overlapping where $d_s$ is the voxel size. Positions of surfels are directly acquired from the sampled point's positions, while the normal vectors are determined by the RANSAC plane fitting to the neighboring points of each sampled point. Plane fitting for points within surfel boundaries is expressed in the plane equation given in Eq.~\ref{eq:plane_equation}. Then the normal vector for surfel $n_s$ becomes the coefficients of the plane equation: 
\begin{equation} \label{eq:plane_equation}
    \begin{aligned}
        Ax + By + Cz + D = 0, \\
        n_s = (n_{sx},n_{sy},n_{sz}) = (A,B,C)
    \end{aligned}
\end{equation} 
$A, B, C, D$ are coefficients of scalar equation of a plane.  $n_s$ becomes normal vector for the plane.

\subsection{Regressing Costs on Local Patches of Terrain} \label{regressing_costs_subsection}
 We regress the cost values for each surfel in the original surfel set $S_O$, and this cost defines a traversability measure for each surfel. We select four \textit{critics} to determine a cost value associated with each surfel:
\begin{itemize}
    \item Tilt of the slope within a surfel, $t_s$.
    \item Average point deviation from the plane of a surfel, $pd_s$, is used to determine roughness.
    \item Max height difference of points within the surfel, $hd_s$.
    \item Ground Clarence to the robot's bottom chassis, $gc_s$.
\end{itemize}
These cost critics aim to determine cost values for terrain patches considering the robot's physical constraints, e.g., roll or pitch angle, terrain roughness, ground clearance, etc.  

The following equations show how these cost critics are computed with the acquired surfel properties. Note that the following cost critics are calculated for each surfel in the surfel set $S_O$.
\begin{equation} \label{eq:critic_equations}
    \begin{aligned}
        t_s = \max(|\arctan(n_{sx},n_{sz})|, |\arctan(n_{sy},n_{sz})|),
    \end{aligned}
\end{equation}
where $\arctan(n_{sx},n_{sz})$ and $\arctan(n_{sy},n_{sz})$ define the roll and pitch of the surfel respectively.
\begin{equation} \label{eq:roughness_equations}
    \begin{aligned}    
        pd_s = 1 / j \sum_{k=1}^{j} |\dfrac{A P_{x} + B P_{y} + C P_{z} + D}{\sqrt{ A^2  + B^2  + C^2 }}|,
    \end{aligned}
\end{equation} 
The average point deviation from the surfel plane $pd_s$ is determined by the sum of all point distances to the surfel plane divided by the number of points within the surfel.
\begin{equation}  
    \begin{aligned}         
        hd_s = \max({P_{z1},..,P_{zj}}) - \min({P_{z1},...,P_{zj}}),
    \end{aligned}
\end{equation} 
The height of a point is its value along the z-axis, with $P_{zj}$ being the z value of $j$th point. $hd_s$ is the maximum height difference among points of the surfel. 
\begin{equation}  
    \begin{aligned}  
        gc_s = \forall k \in \{1,..,j\}, max(\dfrac{A P_{x} + B P_{y} + C P_{z} + D}{\sqrt{ A^2  + B^2  + C^2 }}). 
    \end{aligned}
\end{equation}
Where $gc_s$ is the distance of the farthest point from the surfel plane (along the  normal vector), the $gc_s$ should be less than the distance of the robot chassis to the ground to ensure safe ground clearance. 

The individual critics are then linearly weighted by
\begin{equation} \label{eq:cost_equation}
        J_s = w_t t_s + w_{pd} pd_s + w_{hd} hd_s + w_{gc} gc_s,
\end{equation}
where $w_t$, $w_{pd}$, $w_{hd}$ and $w_{gc}$ are adjustable parameters to weigh each critic according to the desired behavior. In the default configuration, the values for each weight are set to 0.25, meaning they equally contribute to determining the overall traversability cost of a region represented by a surfel.
$J_s$ becomes the cost 
associated with surfel $S_s$. Elevating each surfel
along their normal vector $n_{s}$ by $d_e$ (see Eq.~\ref{eq:elevation_equation}), we achieve a thin layer that follows the structure of terrain with a distance.  $d_e$ is configuration parameter that can be adjusted according to physical properties of robot, distance from robot's center of gravity to the ground is a good default parameter for the $d_e$. 
We refer to this \textit{elevated surfel} set as $S_E$.

\begin{equation} \label{eq:elevation_equation}
       \boldsymbol v_s = \boldsymbol v_s + d_e \boldsymbol n_s,
\end{equation}

\begin{figure}
  \centering
  \includegraphics[width=0.8\linewidth]{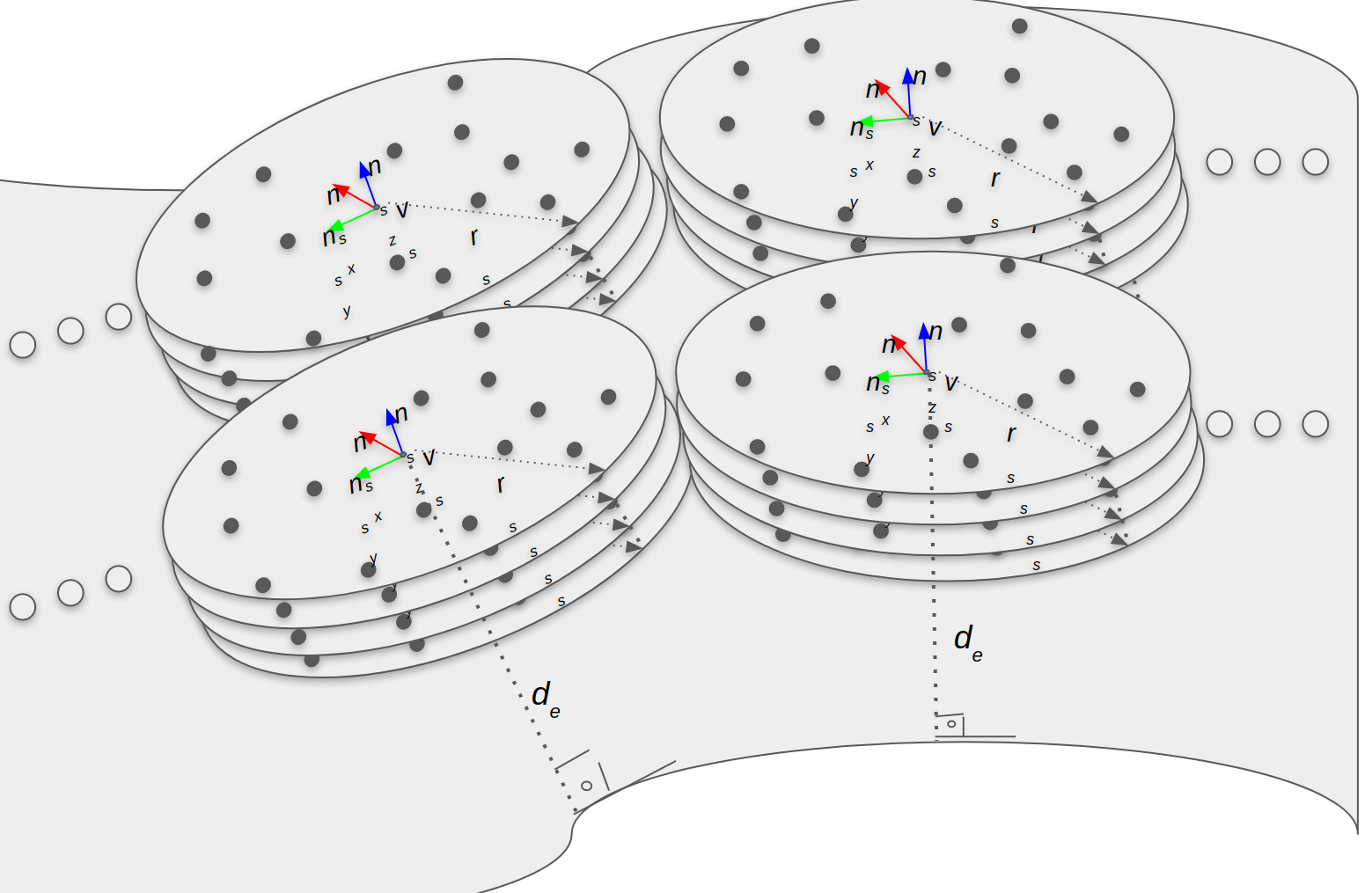}
  \caption{Waffle surfels on top of an unstructured environment patch. Stacked surfels that construct one waffle have identical normal vectors, radius, and cost values. We achieve a constrained state-space suitable for unstructured terrain planning and navigation with the side-by-side collection of surfel waffles.}
  \label{fig:waffle_surfel}
\end{figure}

At this stage, the elevated surfel set $S_E$ resembles a thin layer (consisting of surfels) on top of the terrain underneath. We add a volume to this thin layer by stacking several surfels together (along with their normal vectors) with a step size $s_{size}$. An illustration of this process is depicted in Fig.~\ref{fig:waffle_surfel}. The result of this stacking looks like a ``waffle'' that can be approximated to a cylinder if $s_{size}$ is chosen small enough. This volume is then used for planning and navigation on unstructured terrains. We refer to this volume as elevation state-space.

\begin{figure}
  \centering
  \includegraphics[width=0.9\linewidth]{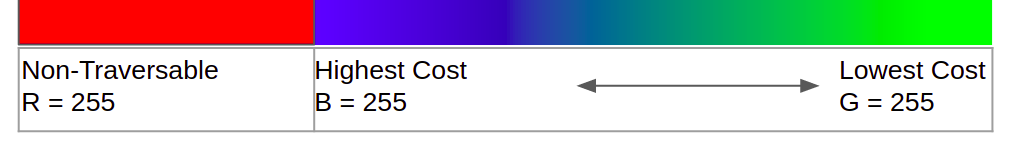}
  \caption{Cost values are encoded with RGB channels to the terrain point cloud map.}
  \label{fig:cost_range}
\end{figure}

\subsection{Integrating the Physical and Kinematic Constraints}

A robot navigating in unstructured terrain is subject to the limitations of maneuvers due to its physical and kinematic constraints.

\begin{figure}
  \centering
  \includegraphics[width=0.8\linewidth]{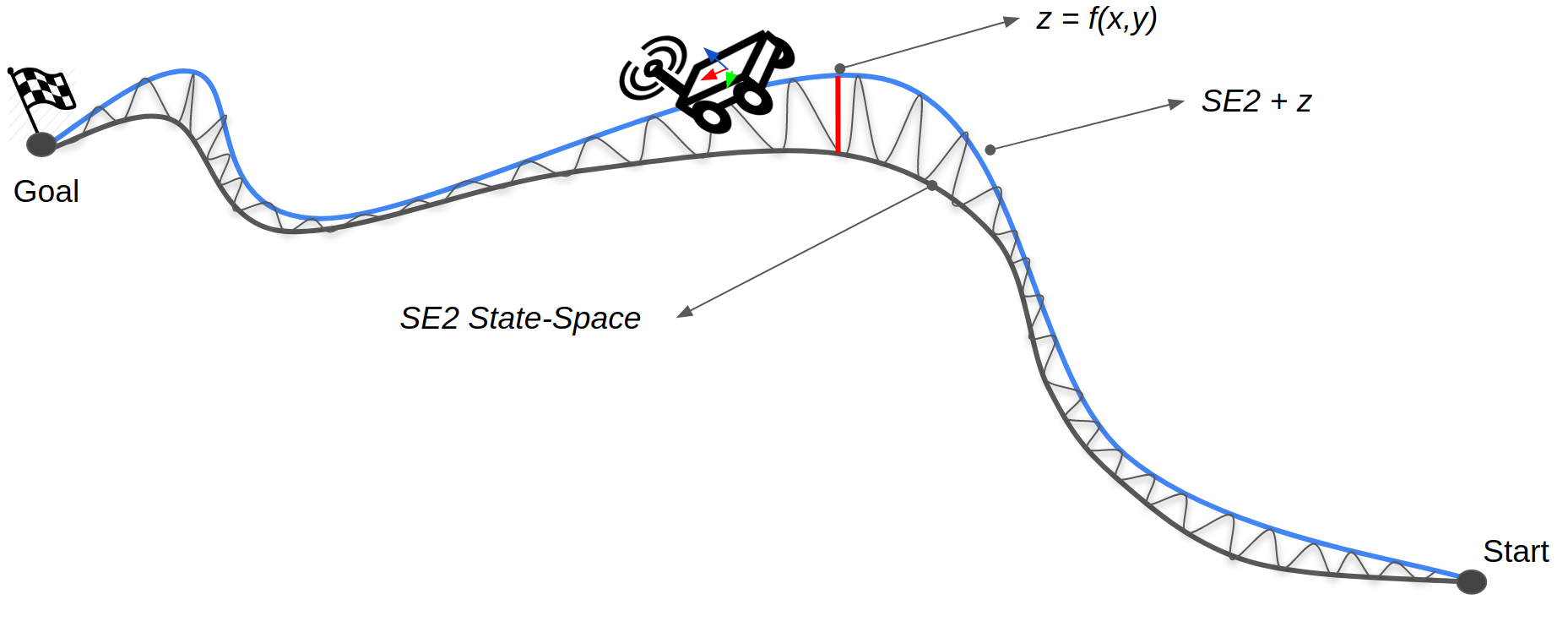}
  \caption{Elevation state-space is composed of SE2 and z components. We account for the kinematic constraints of an Ackerman type by considering Dubins or Reeds-Sheep space for the SE2 element.}
  \label{fig:ee_state_space}
\end{figure}

Physical constraints are robot-specific parameters. For instance, the maximum tilt angle that a robot
can bear while navigating is a robot-specific physical limitation. Another robot-specific limitation is ground clearance~\cite{PAPADAKIS20131373} which ensures contact-free robot chassis-terrain interaction. We address physical constraints by regulating our cost critics (see~Sec.~ \ref{regressing_costs_subsection}) to a configurable range. The range for each cost critic is an adjustable parameter and can be adapted according to the robot type and desired behavior.
For instance, setting a tight range for cost critic $t_s$ (tilt of the terrain) from Sec.~\ref{regressing_costs_subsection} will lead to a more inclination-sensitive traversability map; thus, the robot will not navigate with such settings through highly inclined regions, see Fig.~\ref{fig:surfel_cost}. 
In Fig.~\ref{fig:surfel_cost}, the red-colored patches are marked as non-traversable, resulting in cost critics being out of defined ranges.  
Kinematic constraints, e.g., for the Ackerman type of robot, are considered in the
planning stage. Typically, Dubins~\cite{dubins} or Reeds-Sheep~\cite{reeds-sheep} shortest curves handle these constraints. When the planning problem is aimed at unstructured terrains, the situation gets more complicated as traditionally, Dubins and Reeds-Sheep are computed on a 2D plane, but an additional dimension ($z$) exists for unstructured terrains.
Our method deals with this situation by decomposing SE2 and z as separate components (see  Fig.~\ref{fig:ee_state_space}) of elevation state-space. SE2 state-space is used for robots operating in a flat environment with three degrees of freedom ($x,y, heading$). Dubins or Reeds Sheep variants of SE2 members are used to account for kinematic constraints related to the maneuvers. The constraint of the z component, which enforces constant contact of robot base and terrain, is handled by the elevated surfels introduced in Sec.~\ref{regressing_costs_subsection}

\subsection{Implementation Details} \label{implementation_details}

We implement the new state-space into the OMPL framework. The implementation is done in the C++ programming language.
A few substantial C++ objects that construct the body of our state-space implementation are Elevation State-Space, Surfel-based Valid State Sampler (SVSS), and Surfel Cost Optimization Objective (SCOO).
We can distinguish between traversable and non-traversable regions with the cost regression explained in Sec.~\ref{regressing_costs_subsection}. SVSS exploits this fact and only generates samples within traversable areas.
It is also possible to enable discrete probability distribution to give a higher probability of drawing samples from lower-cost regions. The samples are drawn directly from the elevated surfel set $S_E$, in Eq.~\ref{discrete_dist}, we weigh the likelihood of a sample being drawn according to their costs. The higher a surfel cost is, the less likely \textit{SVSS} will mark it as a sample for OMPL planners, the probability is calculated with \autoref{discrete_dist}.  
\begin{equation} \label{discrete_dist}
\begin{split}
    P(i| (M / J_0) , (M / J_1), ..., (M / J_{n-1})) = \\
   \dfrac{J_i}{\sum_{k=0}^{n-1} M/J_k}, \space \space 0 \leq i < n   
\end{split}
\end{equation}
Where $M$ is the max cost threshold, we assume 255. $P$ describes the probability of surfel $S_i$ being drawn from surfel set $S_E$ as a sample, and $J_i$ is the cost associated with surfel $S_i$. 

Some OMPL planners, such as RRT* and PRM*~\cite{karaman_frazolli} referred to as \textit{optimizing planners}, can use an objective function to optimize a path according to a given objective. The optimizing planners use the objective function's value to judge the quality of the path to improve the quality of the path based on this. \textit{SCOO} sums up costs given in the Eq.~\ref{eq:cost_equation}, \textit{SCOO} can be used with other optimization objectives such as shortest length obstacle clearance objectives already present in OMPL. 

\section{Experiments and Validation} \label{experiments}

We use data from a real environment consisting of unstructured terrain with various objects to evaluate our method. 
We use a LIDAR-based SLAM approach to acquire the natural environment's point cloud map seen in Fig.~\ref{fig:case_0_result}.
Recent SLAM approaches~\cite{liosam2020shan, legoloam2018shan} and~\cite{lee} can produce dense and accurate point cloud maps using a LIDAR, IMU, and optionally GNSS.
LIO-SAM by Shan et al.~\cite{liosam2020shan} proved to produce reliable maps of these methods. Therefore, we run our tests on the maps produced by LIO-SAM. In addition to actual data from the real world, we use point cloud maps generated from the Gazebo simulator. As our robot platform,  Thorvald II by Grimstad et al.~\cite{GRIMSTAD20174588}, Thorvald II is a modular agricultural robot suitable for unstructured environment navigation. 



\begin{figure*}
  \centering
  \includegraphics[width=1.0\linewidth]{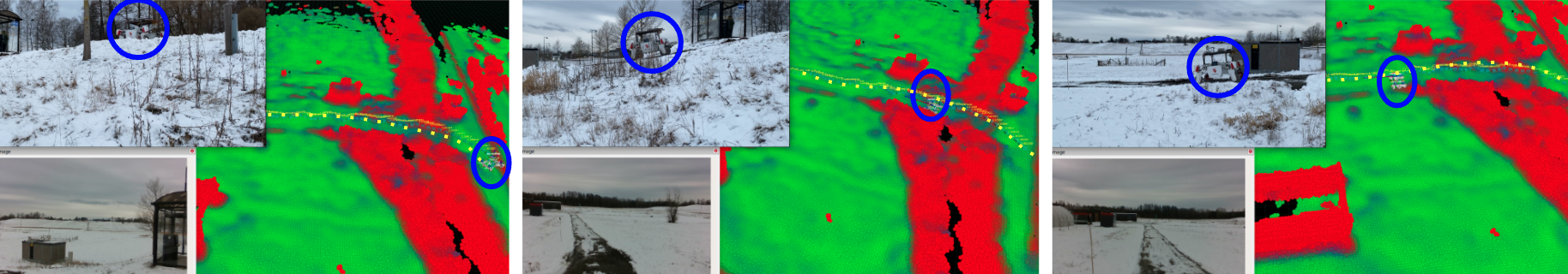}
  \caption{A sequence of images showing auto navigation of Thorvald II robot in a real unstructured snowy terrain with the help of \textit{elevation} state-space. In sequences, at the top-left third-person view of the robot, at the bottom-left robot's POV, and on the right side, the RVIZ view of the map is depicted. In the RVIZ view, the planned path also can be seen(yellow-colored line).}
  \label{fig:seq}
\end{figure*}

\subsection{Comparison to Standard ROS Navigation Framework}\label{nav2_clash}

\begin{figure*}[!htb]%
    \centering
    \subfloat[\centering]{{\includegraphics[width=18cm]{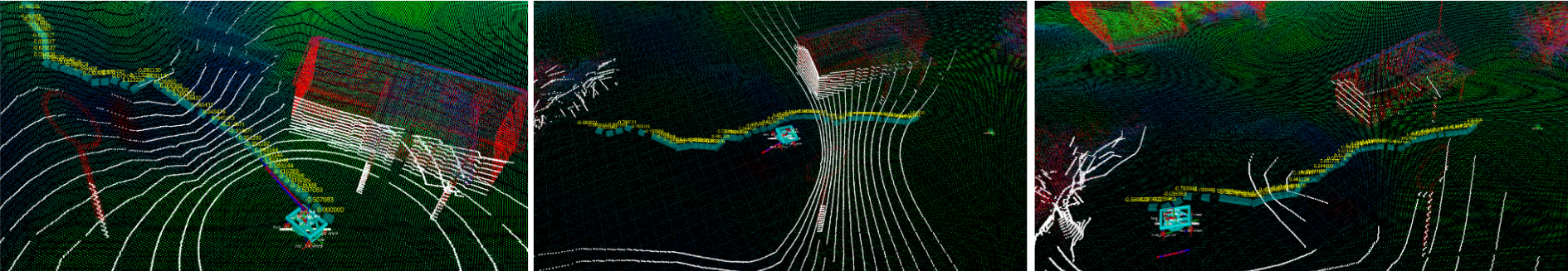} }}%
    \qquad
    \subfloat[\centering]{{\includegraphics[width=18cm]{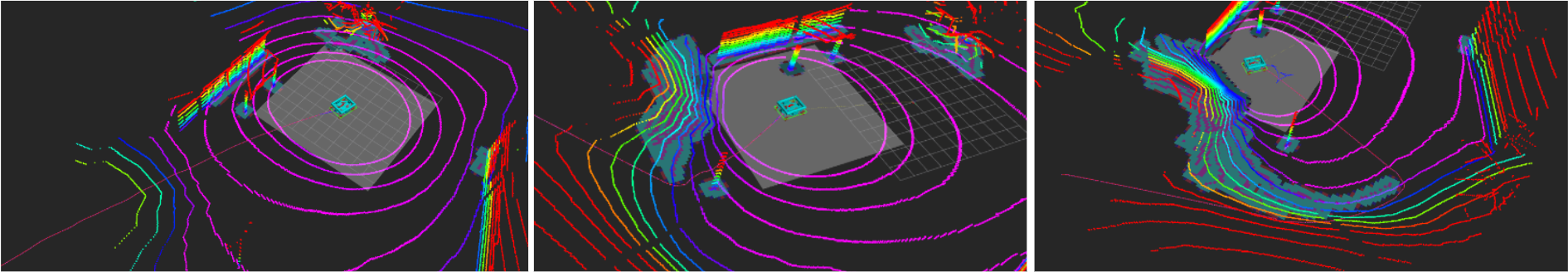} }}%
   \caption{The robot effortlessly goes through an inclined hill and reaches the desired goal in elevation state-space (a). In the classical navigation framework, as the robot gets closer to the inclined terrain, it perceives the hill as an \textit{obstacle}. While in reality, it is traversable for the given robot model. This leads to confusion on the planner and controller, ultimately causing navigation failure (b).}%
    \label{fig:nav2_clash_fig}%
\end{figure*}

Two approaches, Mesh Nav~\cite{mesh_nav} and Navigation2~\cite{macenski2020marathon2}, have organized software implementations that we may compare our approach as baselines. Navigation2 is a widely used navigation framework that is built on top of a well-matured software stack~\cite{tbd}. The primary use of this framework is in indoor scenes; however, outdoor scenes are also possible. We realize that, for even terrain patches, both our method and Navigation2 perform as expected. The performance difference reveals itself when the terrain is uneven (e.g., steep inclinations). Our approach correctly models the unstructured nature of outdoor scenes, leading to superior performance over Navigation2. An example case is depicted in Fig.~\ref{fig:nav2_clash_fig} where a robot (Ackerman type) is given an identical navigation task over an unstructured terrain. Thanks to the terrain feature-aware nature of elevation state-space, the planning, and navigation tasks are effortlessly handled, whereas Navigation2 fails to address the tasks due to a lack of accurate representation of uneven terrain since it bases on a 2D occupancy grid. 

\subsection{Edge-Case Tests in Real Environment} \label{rigorous_testing}
The edge cases are aimed to qualitatively assess the planning and navigation performance in challenging parts of an actual environment map. In the first case, the robot must navigate from \textit{various} poses to one goal pose in the map depicted in Fig.~\ref{fig:case_0_result}.   
\subsubsection{Navigating through an unstructured inclination} \label{case_0}
\begin{figure}
  \centering
  \includegraphics[width=1\linewidth]{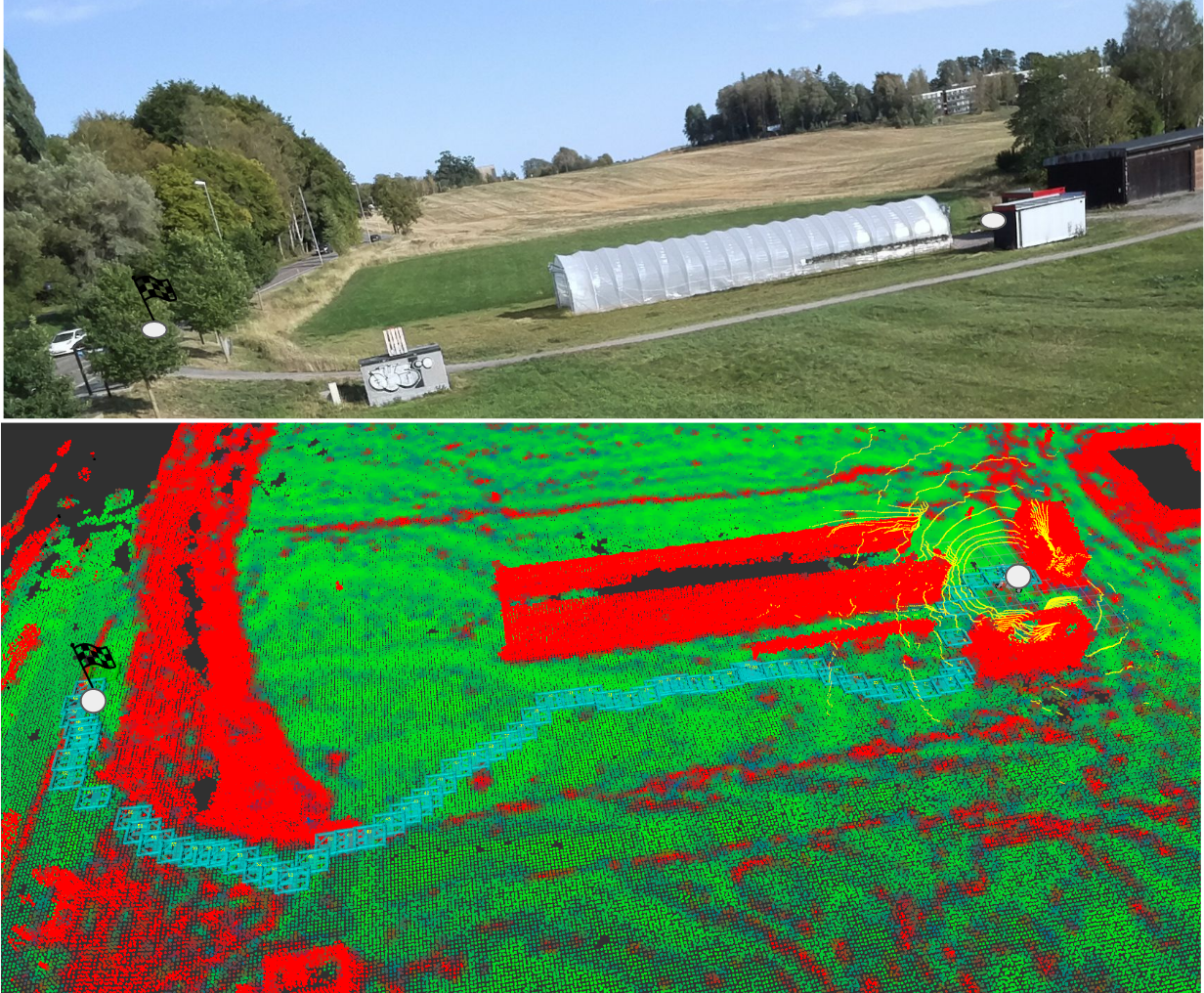}
  \caption{Example result of navigating through an unstructured inclination mentioned in Sec.~\ref{case_0}. The robot start position is marked with white circle and the white circle with the flag is the goal position, in the upper side, a drone image of the case is presented. At the bottom, a cost regressed map of the environment and produced path (in cyan) are visualized.}
  \label{fig:case_0_result}
\end{figure}

In this case, the robot is expected to navigate through an unstructured inclination; see Fig.~\ref{fig:case_0_result}. Although we assume a robot model that can take on unstructured terrain, the robot is expected to follow a smoother and less inclined path to minimize energy consumption and risk factors. For this case, there are approximately 2 meters of elevation variance from start to goal, with the total distance to the goal being around 65 meters.

The planning and consecutive navigation for this case consistently resulted in success. See Fig.~\ref{fig:case_0_result} Fig.~\ref{fig:seq} for instance. The planned paths are steered towards feasible regions, mainly when the inclination raises. Fig.~\ref{fig:case_0_result} and Fig.~\ref{fig:seq} belong to the same scene, but experiments in Fig.~\ref{fig:seq} were performed during winter. 

\subsubsection{Navigating through the excavation of the building site} \label{case_1}
In this case, the robot tries to navigate through an excavation residual of a building in the real environment. 
The challenge of this case was the height difference of the excavation area from the rest of the ground plane and its non-traversable edges formed by piles of sand and gravel resulting from some building works. This limits feasible regions for the safe navigation of the robot. Repetitive planning and navigation queries from the different start positions to the goal inside the excavation resulted in successful navigation.


\subsection{Comparison of Planners Performance in Elevation State-space} \label{problem_generation}

\begin{figure}
  \centering
  \includegraphics[width=0.8\linewidth]{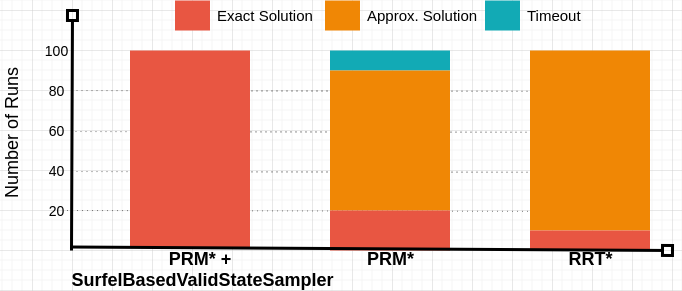}
  \caption{Success rates from OMPL planners for randomly generated 100 planning problems on an unstructured terrain. Without exploiting surface information of elevation,  planners suffer as most sampled states are generated in undesired areas. However, guided by \textit{SVSS}, the first instance of PRM* solves all planning problems.}
  \label{fig:benchmark_result}
\end{figure}

The selected edge cases are insufficient to draw general conclusions on planners' performance in the proposed state-space. We perform benchmarking for the success rates of RRT* and PRM* planners from the OMPL on a randomly generated problem set to draw a general conclusion on planners' performance in the elevation state-space. The randomly generated set consists of 100 planning problems in the real or simulated map (randomly chosen for each problem) used in the edge-case tests. All randomly generated problems satisfy the following conditions:
\begin{itemize}
    \item The Euclidean distance between the randomly generated start and goal pose is from 40 meters up to 100 meters.
    \item There exists at least one exact solution for the generated problem. 
\end{itemize}
We verify that at least one solution exists for the generated problem by setting the timeout parameter to a very long wait time, 5 min. If the planner found an exact solution within that time interval, we add the generated problem for benchmarking; if the planner found no solution for the generated problem, we skip it and generate another one until 100 problems are available for benchmarking. A solution is considered exact if the path has been connected to the goal state with less than 0.2 meters gap and approximate if this gap is less than 10 meters. The success rate results from 100 runs are given in Fig.~\ref{fig:benchmark_result}.


\subsubsection{\textit{SCOO} assessment} \label{cost_optimization}
\begin{figure}
  \centering
  \includegraphics[width=1.0\linewidth]{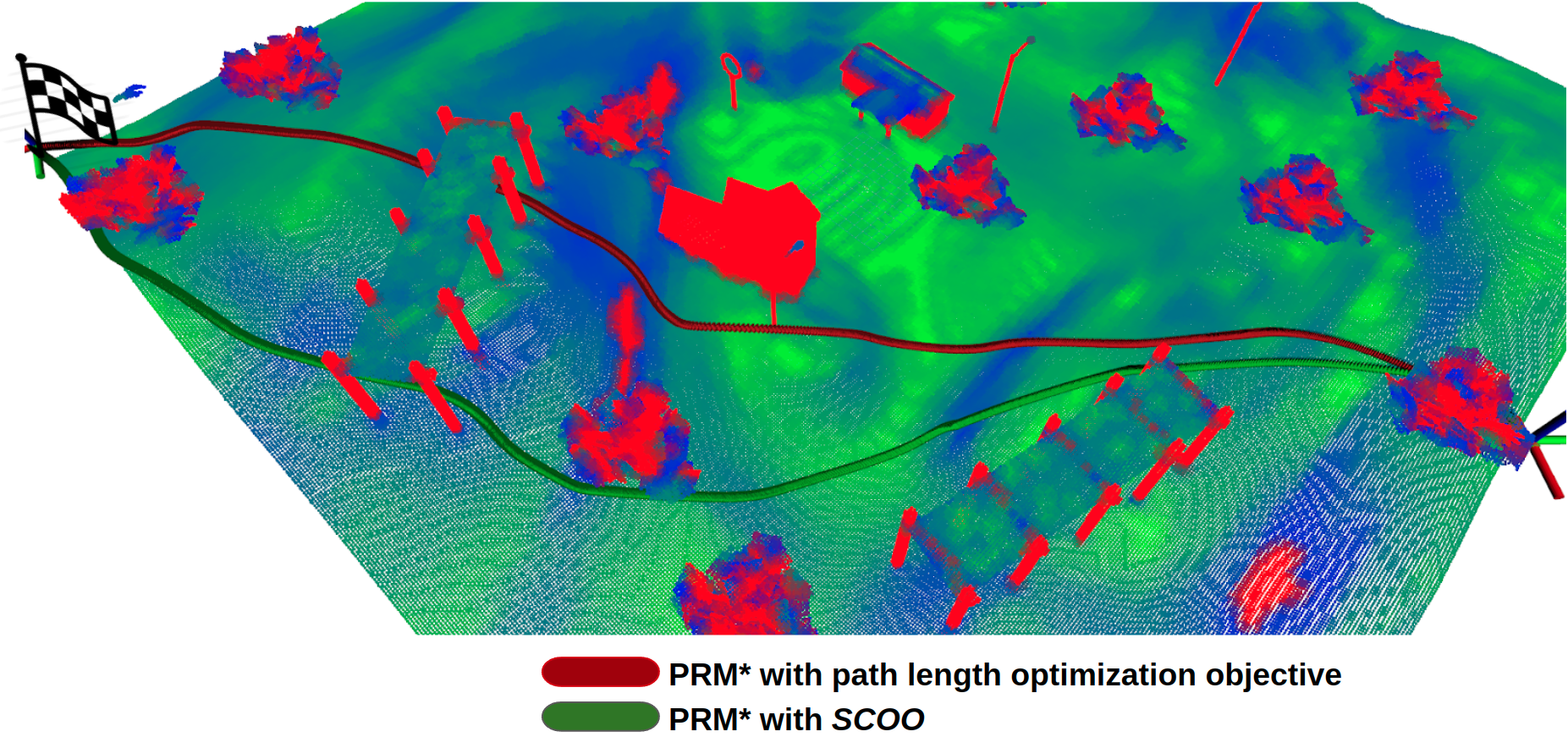}
  \caption{
  Compared to the path length optimization objective, \textit{SCOO} maintains approximately the same path length with lower traversability cost overall, improving the safety and energy consumption during navigation. Refer to Fig.~\ref{fig:cost_range} for the relation between RGB color and cost values. \textit{SCOO} can also be used with other optimization objectives to achieve composite behavior.}
  \label{fig:surfel_cost}
\end{figure}

We introduced \textit{SCOO} in Sec.~\ref{implementation_details}. It is worth noting how this optimization objective affects the behavior of planners. Planners tend to avoid higher-cost areas with SCOO, including steep inclinations and rougher terrain patches. In the figure, we compare results from two instances of PRM* where one instance is configured with SCOO, and the other instance is configured with the length optimization objective. The length optimization objective tries to maintain the shortest path possible. In contrast, the SCOO considers the integrated cost values along this path, thanks to our state-space representation's available traversability measures(surfel costs). In this case, the SCOO gets safer and less energy-consuming paths. See an example in Fig.~\ref{fig:surfel_cost}, where the SCOO steers the path towards greener areas, indicating lower-cost zones.

\subsection{Discussion} \label{discussion}

We provided results from randomly generated problems in Sec.~\ref{problem_generation} to evaluate the use and benefits of the proposed method based on the robot model on hand. 
Point clouds are accurate approximations of the actual 3D world and are the most common output format from 3D SLAM methods. Therefore, we argue that it is beneficial to base robot navigation directly on raw point clouds. We used the intermediate surfel representation since we are interested in leveraging the abstract features of point cloud surfaces. 
Concepts that proposed method introduced, i.e., \textit{SVSS} and \textit{SCOO}
proved to compliment the performance of the OMPL planners in the context of unstructured terrain planning see Fig.~\ref{fig:benchmark_result}. 

With the existing state-space implementations in OMPL, such as SE3 SE2, it was impossible to achieve feasible path plans in unstructured terrains. The simple reason for this is that these state-spaces are not aware of constraints(kinematic or physical) that the planner must respect. In contrast, these constraints are considered and respected in the proposed elevation state-space. With the introduced cost critics, we can readjust the behaviors of planners according to the desired result. For instance, increasing the weight of average point deviation from the surfel plane critic ($pd_s$ in Sec.~\ref{regressing_costs_subsection}) forced plans to be generated in smoother areas rather than short-cutting towards the goal. In unstructured terrains, defining the characteristics of an optimal path is challenging. For instance, which path is the most optimal and desired path with minor energy consumption, the shortest length, or the least risk of rollover?. We argue that the answer is application-specific. Our method's concept of incorporating traversability costs to planning through \textit{SCOO} becomes helpful in defining what should be the characteristics of an optimal path because it is possible to weigh the cost critics. 

\section{Conclusion} \label{conclusion}
This paper used the surfel representation of underlying point clouds to introduce a constraint-aware state-space. The outcomes of the experiments show that the method can be used to navigate in difficult, uneven surroundings.
Within the suggested method, it is also possible to account for the kinematic and physical constraints of different robot kinds. The inefficiency of our approach when dealing with kinematic constraints is one of its flaws. When employing Dubins or Reeds Sheep state-spaces, it takes a long time for planners to converge to near-optimal paths. In the future, we'd like to expand on this work by looking at efficiency difficulties while accounting for kinematic constraints and standardizing motion planning and navigation for unstructured terrains using point clouds.
\bibliographystyle{abbrv}
\bibliography{root}
\end{document}